\documentclass[10pt,twocolumn,letterpaper]{article}

\usepackage[pagenumbers]{wacv} 

\usepackage{graphicx}
\usepackage{optidef}
\usepackage{amsmath}
\usepackage{amssymb}
\usepackage{booktabs}
\usepackage{stix}
\usepackage{amsmath,mleftright,mathtools}
\usepackage{multirow}
\usepackage{hhline}
\usepackage{caption}
\usepackage{makecell}
\usepackage{pifont}
\usepackage{subcaption} 
\usepackage[export]{adjustbox}
\usepackage{dblfloatfix} 
\usepackage{enumitem}

\usepackage[pagebackref,breaklinks,colorlinks]{hyperref}

\usepackage[capitalize]{cleveref}
\crefname{section}{Sec.}{Secs.}
\Crefname{section}{Section}{Sections}
\Crefname{table}{Table}{Tables}
\crefname{table}{Tab.}{Tabs.}
\DeclareMathAlphabet\mathbfcal{LS2}{stixcal}{b}{n}
\DeclareMathOperator*{\argmin}{argmin}
\def\wacvPaperID{1013} 
\def\confName{WACV}
\def\confYear{2025}

\begin{document}

\title{Deep Joint Unrolling for Deblurring and Low-Light Image Enhancement (JUDE)}

\author{
    Tu Vo \,\,\,\, Chan Y. Park \\
    KC Machine Learning Lab \\
    {\tt\small \{tuvv, chan.y.park\}@kc-ml2.com}
}
\maketitle

\begin{abstract}
   Low-light and blurring issues are prevalent when capturing photos at night, often due to the use of long exposure to address dim environments. Addressing these joint problems can be challenging and error-prone if an end-to-end model is trained without incorporating an appropriate physical model. In this paper, we introduce \textbf{JUDE}, a Deep \textbf{J}oint \textbf{U}nrolling for \textbf{D}eblurring and Low-Light Image \textbf{E}nhancement, inspired by the image physical model. Based on Retinex theory and the blurring model, the low-light blurry input is iteratively deblurred and decomposed, producing sharp low-light reflectance and illuminance through an unrolling mechanism. Additionally, we incorporate various modules to estimate the initial blur kernel, enhance brightness, and eliminate noise in the final image. Comprehensive experiments on LOL-Blur and Real-LOL-Blur demonstrate that our method outperforms existing techniques both quantitatively and qualitatively.
\end{abstract}

\section{Introduction}
\label{sec:intro}
Photographs taken in dim lighting often employ long exposure methods to sufficiently illuminate the scene. Unfortunately, long exposures tend to cause motion blur due to camera movement and dynamic elements within the frame. As a result, night photography frequently suffers from both poor lighting and blurriness. These issues combined can obscure image clarity, causing color and texture distortions that adversely affect advanced vision tasks such as image classification and object detection. Hence, the simultaneous enhancement of low-light conditions and image deblurring has become a crucial area of interest, attracting significant focus in image processing and vision research.

Earlier approaches have treated low-light enhancement \cite{zhang2021beyond, guo2020zero, wei2018deep, li2021low, wang2023fourllie, retinexformer, yang2021sparse, wang2023ultra} and image deblurring \cite{Cho2021RethinkingCA, fftformer, sharpformer, nah2017deep, tao2018scale} as separate issues. Each technique assumes distinct conditions for each challenge, resulting in a lack of cohesive solutions that effectively tackle both visibility and sharpness issues together. Low-light enhancement techniques concentrate on correcting illumination and reducing noise, but they overlook the problem of spatial blurring. Conversely, existing deblurring methods are typically developed and trained on daytime images, which often makes them ineffective for nighttime scenes. 


Zhou \textit{et al.} \cite{lednet} developed an end-to-end network by training on a specialized dataset designed for both low-light enhancement and deblurring. This combined approach tackles noise and blurring simultaneously, unlike previous methods which focused on one problem at a time using an encoder-decoder architecture. Hoang \textit{et al.} \cite{feli} introduced FELI which utilizes a trainable Decomposer grounded in Retinex theory \cite{guo2016lime}, training with a customized contrastive regularization term in a knowledge distillation manner. While these models effectively address the limitations of tackling these issues independently, treating a deep learning model as a black box makes it challenging to analyze the behavior of CNNs. Moreover, deep learning-based models tend to suffer from data dependency, which restricts their generalization capabilities.

Algorithm unrolling \cite{Monga2019AlgorithmUI, eldar}, a recently developed technique, bridges model-based and learning-based approaches by integrating iterations of model-based algorithms as layers in deep networks. This method enhances interpretability and theoretical grounding by leveraging mathematical models. Due to its strong performance and generalization capabilities, algorithm unrolling has been applied to various image processing tasks \cite{Monga2019AlgorithmUI, liu2021ruas, Chen, usrnet, eldar}. In this paper, we propose a Deep \textbf{J}oint \textbf{U}nrolling For \textbf{D}eblurring and Low-Light Image \textbf{E}nhancement (\textbf{JUDE}) which employs unrolling techniques to numerically formulate the Low-Light Image Enhancement and Deblurring problem, incorporating two data terms: one for image deblurring and another for image decomposition to sharpen and decompose the low-light blurry images in RGB color space. To enhance robustness, regularizers are included in this formulation, which is then iteratively solved using the Augmented Lagrange Multiplier (ALM) method \cite{lin2010augmented} with an unrolling strategy. The optimization variables and regularizers are updated in each iteration through closed-form solutions and learned deep networks, respectively, resulting in sharp low-light reflectance and illuminance. Finally, we implement illuminance adjustment and reflectance denoiser modules that faithfully enhance the illumination map and remove possible noise, generating non-blurry bright images. 

In summary, this paper makes three main contributions:
\begin{itemize}[noitemsep]
    \item Drawing from traditional model-based approaches, we introduce JUDE, a novel deep joint unfolding network for low-light image enhancement and deblurring. JUDE comprises four modules: blur kernel estimation, optimization, illumination enhancement, and reflectance denoiser. 
    
    \item The optimization module within JUDE translates the optimization process into a deep network, harnessing the robust modeling capabilities of learning-based methods to incorporate data-driven priors adaptively.
    \item We conducted experiments using the LOL-Blur \cite{lednet} and Real-LOL-Blur \cite{rim2020real} datasets. Both quantitative metrics and qualitative results highlight the superior effectiveness of our approach in both synthetic and real-world datasets.

\end{itemize}

\section{Related Works}
\subsection{Low Light Image Enhancement}
Low-light image enhancement (LLE) techniques focus on improving the visual quality of underexposed images. Traditional methods include Lee et al.'s approach \cite{lee}, which employs a tree-like layered structure to represent gray-level differences and solves a constrained optimization problem to determine the transformation function. With the rise of deep learning techniques, learning-based LLE models are now taking dominance. Their methods range from autoencoder \cite{lore2017llnet}, multi-stage enhancement \cite{makwana2024livenet}, 3D lookup tables \cite{yang2022adaint}, generative models \cite{yixunpeng}, and Retinex-based decomposition \cite{zheng, wei2018deep, zhang2019kindling, yang2021sparse, wang2019underexposed, retinexformer, xuke, liu2021ruas}. Retinex theory has been proven to be effective and widely used in LLE problems as it effectively decomposes the image into two separate parts for further processes.



\subsection{Image Deblurring}
Image deblurring techniques restore sharpness to blurred images through various methods. With advancements in deep learning, image deblurring has seen significant improvements. Cho \textit{et al.}\cite{Cho2021RethinkingCA} introduced a network with multi-scale outputs for comprehensive supervision, resulting in low computational overhead. Tu \textit{et al.}\cite{tu2022maxim} proposed MLP-based network blocks that required larger batch sizes during training, while Mao \textit{et al.}\cite{mao2023intriguing} developed efficient network blocks using selective FFT operations in the frequency domain. In addition, by taking advantage of transformers, Chen et al. \cite{fftformer} employed the Diffusion model in a compact latent space to generate prior features for the deblurring process, implemented through a regression-based method, Yan et al. \cite{sharpformer} directed their efforts towards preserving local details by utilizing Transformer blocks, and Kong et al. \cite{kong2023efficient} introduced an innovative application of self-attention mechanisms, extending them from the spatial domain to the frequency domain. Among these, a subset of image deblurring focuses on scenarios where blurry images are captured under low-light conditions \cite{huzhe, zhouchu, zhou2023deblurring, chenliang}. However, these works' primary objective is to deblur the captured scenes without addressing any light level adjustments.
\vspace{-2.5mm}

\subsection{Joint Degradation Methods}
Many existing methods in image enhancement and restoration typically address singular tasks like denoising, deblurring, or super-resolution. Yet, real-world scenarios often present images with multiple degradations. Recently, there has been a trend towards developing integrated solutions that tackle these simultaneous challenges. For instance, Xing \textit{et al.} \cite{xing2021end} introduced a comprehensive approach that combines demosaicing, denoising, and super-resolution tasks within a stacked module network structure, optimizing color extraction, feature processing, and image reconstruction. Yang \textit{et al.} \cite{yang2021srdn} proposed a unified network specializing in joint deblurring and super-resolution for space targets, while Lu \textit{et al.} \cite{lu2022progressive} designed a framework with dual branches dedicated to enhancing low-light conditions and removing noise from Raw images simultaneously. 


For low-light image enhancement and deblurring, efforts by Zhou \textit{et al.} \cite{lednet} marked the first attempt to concurrently address low-light enhancement and deblurring using an encoder-decoder model. Followed by Hoang \textit{et al.} \cite{feli}, FELI is a Retinex-inspired learnable Decomposer for scene recovery and enhanced feature encoding with blur-aware input reconstruction. Although these methods can tackle the joint problem, their effectiveness notably depends on the training data's diversity \cite{Kim_2019_CVPR}, leading to limited generalization capability. Furthermore, their behavior is difficult to interpret and analyze due to the processing of features in high-dimensional feature spaces. 

Recently, a novel technique called algorithm unrolling \cite{Monga2019AlgorithmUI} has been developed to address these limitations \cite{Monga2019AlgorithmUI, liu2021ruas, Chen}. This method reformulates the iterations of an optimization algorithm as CNN layers. In this paper, we leverage algorithm unrolling to introduce an unrolled deep network that incorporates physical models for both blurring and low-light conditions, effectively merging the advantages of both model-based and learning-based methods. 


\section{Proposed Method}
Figure \ref{fig:jude} presents the overall architecture of the proposed unrolling algorithm. To address the low-light blurring problem, in Section \ref{optimization}, we first formulate an optimization problem to deblur the input image and decompose it into illuminance and reflectance components and explain the solution in Section \ref{solution}. The decomposed illuminance is subsequently enhanced using an Illuminance Enhancer, while the reflectance undergoes additional denoising before being recombined to produce a non-blurry, well-lit image. These components will be discussed in Section \ref{network}, along with the Kernel Estimation.

\subsection{Optimization Problem Formulation}
\label{optimization}

We formulate the joint Retinex decomposition and deblurring problem as follows:

\begin{mini}|l|
{\mathbfcal{I},\mathbfcal{H},\mathbfcal{R},\mathbfcal{L}}{\frac{\lambda_{1}}{2}\| \mathbfcal{X} - \mathbfcal{H}\otimes \mathbfcal{I}\|_{2}^{2}
+\frac{\lambda_{2}}{2}\|\mathbfcal{I} - \mathbfcal{R}\odot\mathbfcal{L} \ \|_{2}^{2}}
{}{}
\label{eq:original}
\end{mini}

where \(\mathbfcal{X}, \mathbfcal{I}, \mathbfcal{H}, \mathbfcal{R}, \mathbfcal{L}\) are low-light blurry image, low-light non-blurry image, blur kernel, reflectance, and illuminance of the low-light non-blurry image, respectively. \(\odot\), \(\otimes\), \(\lambda_{1}\), and  \(\lambda_{2}\) denote the element-wise multiplication, the convolution operation, and balancing parameters, respectively. 

In Equation \ref{eq:original}, the first term formulates the deblurring model, while the second term corresponds to the Retinex-based decomposition. Furthermore, Equation \ref{eq:original} also assumes that a low-light blurry image \(\mathbfcal{X}\) should be deblurred into \(\mathbfcal{I}\) before being decomposed into \(\mathbfcal{R}\) and \(\mathbfcal{L}\). These assumptions are ill-posed and might be incorrect, therefore we add three more regularizers to compensate for these assumptions, as shown in the equation \ref{eq:original_full} below:

\begin{mini}|l|
{\mathbfcal{I},\mathbfcal{H},\mathbfcal{R},\mathbfcal{L}}{\frac{\lambda_{1}}{2}\|\mathbfcal{H}\otimes \mathbfcal{I} - \mathbfcal{X} \|_{2}^{2} + \frac{\lambda_{2}}{2}\|\mathbfcal{I} - \mathbfcal{R}\odot\mathbfcal{L} \ \|_{2}^{2}}{}{} 
\breakObjective{ + \Phi_{Y}(\mathbfcal{I}) + \Phi_{R}(\mathbfcal{R}) + \Phi_{L}(\mathbfcal{L})} 
\label{eq:original_full}
\end{mini}

\subsection{Solution to the Optimization} \label{solution}
We apply the Augmented Lagrangian Method (ALM) \cite{lin2010augmented} to iteratively solve the optimization problem in Equation (\ref{eq:original_full}). To streamline this process, we first restructure Equation (\ref{eq:original_full}) for variable splitting, introducing auxiliary variables \(\mathbfcal{P}, \mathbfcal{Q}\), and \(\mathbfcal{Z}\).

\begin{mini}|l|
{\mathbfcal{I},\mathbfcal{H},\mathbfcal{R},\mathbfcal{L}}{\frac{\lambda_{1}}{2}\|\mathbfcal{H}\otimes \mathbfcal{I} - \mathbfcal{X} \|_{2}^{2} + \frac{\lambda_{2}}{2}\|\mathbfcal{Z} - \mathbfcal{P}\odot\mathbfcal{Q} \ \|_{2}^{2}}{}{} 
\breakObjective{ + g_{Z}(\mathbfcal{Z}) + g_{P}(\mathbfcal{P}) + g_{Q}(\mathbfcal{Q})} 
{}{}
\addConstraint{\mathbfcal{P}=\mathbfcal{R}}
\addConstraint{\mathbfcal{Q}=\mathbfcal{L}}{}
\addConstraint{\mathbfcal{Z}=\mathbfcal{I}}{}
\label{eq:withauxies}
\end{mini}

The ALM method solves a sequence of unconstrained sub-problems derived from an original constrained optimization problem. In particular, we define the augmented Lagrangian function \(\mathbfcal{E}\) for the problem outlined in Equation (\ref{eq:withauxies}) as follows:

\begin{equation}
  \begin{aligned}
    \mathbfcal{E}(\mathbfcal{I},\mathbfcal{H},\mathbfcal{R},\mathbfcal{L},\mathbfcal{Z},\mathbfcal{P},\mathbfcal{Q}) = \frac{\lambda_{1}}{2}\|\mathbfcal{H}\otimes \mathbfcal{I} - \mathbfcal{X} \|_{2}^{2} \\ 
          + \frac{\lambda_{2}}{2}\|\mathbfcal{Z} - \mathbfcal{P}\odot\mathbfcal{Q} \|_{2}^{2} + g_{Z}(\mathbfcal{Z}) + g_{P}(\mathbfcal{P}) + g_{Q}(\mathbfcal{Q}) \\
          + \langle \mathbf{\mathbf{\Gamma}}, \mathbfcal{R} - \mathbfcal{P}\rangle + \frac{\lambda_{3}}{2}\|\mathbfcal{R} - \mathbfcal{P}\|_{2}^{2} \\
         + \langle \mathbf{\mathbf{\Omega}}, \mathbfcal{L} - \mathbfcal{Q}\rangle + \frac{\lambda_{4}}{2}\|\mathbfcal{L} - \mathbfcal{Q}\|_{2}^{2} \\
         + \langle \mathbf{\mathbf{\Delta}}, \mathbfcal{I} - \mathbfcal{Z}\rangle + \frac{\lambda_{5}}{2}\|\mathbfcal{I} - \mathbfcal{Z}\|_{2}^{2}
  \end{aligned}
  \label{eq:fullequation}
\end{equation}

where \(\lambda_{1}, \lambda_{2}\, \lambda_{3}\), \(\lambda_{4}\), and \(\lambda_{5}\) are hyper-parameters, \(\mathbf{\mathbf{\Gamma}}, \mathbf{\mathbf{\Omega}}\) and \(\mathbf{\mathbf{\Delta}}\) are Lagrange multipliers, and \(\langle.,.\rangle\) denotes the inner product of two tensors.

Finally, solutions to the optimization problem outlined in Equation (\ref{eq:withauxies}) can be attained by minimizing the augmented Lagrangian function \(\mathbfcal{E}\) presented in Equation (\ref{eq:fullequation}), \textit{i.e.},

\begin{equation}
(\mathbfcal{I}^{*},\mathbfcal{H}^{*},\mathbfcal{R}^{*},\mathbfcal{L}^{*}) = \argmin_{\mathbfcal{I},\mathbfcal{H},\mathbfcal{R},\mathbfcal{L}} \mathbfcal{E}(\mathbfcal{I},\mathbfcal{H},\mathbfcal{R},\mathbfcal{L},\mathbfcal{Z},\mathbfcal{P},\mathbfcal{Q}, \mathbf{\Gamma}, \mathbf{\Omega}, \mathbf{\Delta})
\label{eq:shortequation}
\end{equation}

Dealing with a joint optimization problem, as depicted in Equation (\ref{eq:shortequation}), is not practically feasible. Consequently, this paper adopts the alternating direction method of multipliers \cite{lin2010augmented}. This strategy entails breaking down the optimization into sub-problems concerning variables \(\mathbfcal{I}, \mathbfcal{H}, \mathcal{R}, \mathbfcal{L}\), and multipliers \(\mathbf{\Gamma}, \mathbf{\Omega}, \mathbf{\Delta}\). These sub-problems are solved iteratively and individually, as elucidated below: \\


\textbf{Update} \(\mathbfcal{P}_{t+1}\): At the \textit{t}th iteration, we update \(\mathbfcal{P}\) as
\nolinebreak
\begin{equation}
\begin{aligned}
    \mathbfcal{P}_{t+1} = g_{P}(\mathbfcal{P}) + \argmin_{\mathbfcal{V}}\frac{\lambda_{2}}{2}\|\mathbfcal{Z}_{t} + \mathbfcal{P} \odot \mathbfcal{Q}_{t}\|_{2}^{2} \\
    + \langle\mathbf{\Gamma}_{t}, \mathbfcal{R}_{t}-\mathbfcal{P}\rangle 
    + \frac{\lambda_{3}}{2}\|\mathbfcal{R}_{t}-\mathbfcal{P}\|_{2}^{2}
\end{aligned}
\end{equation} 

\begin{equation}
   = \argmin_{\mathbfcal{P}} g_{P}(\mathbfcal{P}) + \frac{\mathbf{\mathbfcal{Q}_{t}^{2}\lambda_{2} + \lambda}_{3}}{2}\Big(\mathbfcal{P}-\frac{\lambda_{2}\mathbfcal{Z}_{t}\mathbfcal{Q}_{t} + \lambda_{3}\mathbfcal{R}_{t}+\mathbf{\Gamma}_{t}}{\mathbfcal{Q}^{2}_{t}\lambda_{2} + \lambda_{3}}\Big)^{2}
\end{equation}

\begin{equation}
   = \mathbfcal{D}_{g_{P}}(\mathbfcal{P} - \frac{\mathbf{\Psi}_{t}}{\mathbfcal{Q}^{2}_{t}\lambda_{2} + \lambda_{3}})
\end{equation} 

where: \(\mathbf{\Psi}_{t} = \lambda_{2}\mathbfcal{Z}_{t}\mathbfcal{Q}_{t} + \lambda_{3}\mathbfcal{R}_{t}+\mathbf{\Gamma}_{t}\).

The operator \(\mathbfcal{D}_{g_{P}}\) is associated with regularization functions \(g\)(·). Traditionally, these functions were defined by observing specific phenomena in various applications, such as gradient fidelity in image defogging \cite{van2021single}, smoothness of illumination maps for low-light image enhancement \cite{ren2020lr3m}, and sparsity emphasis in image denoising \cite{giryes2014sparsity}. However, relying solely on this observational method can lead to inaccuracies and limit the model's generalization capabilities. Recognizing that these regularizers are more complex and should be represented and learned from real-world data, we use ResUNet \cite{diakogiannis2020resunet} to implement the Data Operator \(\mathbfcal{D}\). This approach allows the model to learn from training data and effectively reconstruct intricate and diverse visual features. During training, the network parameters of \(\mathbfcal{D}_{\mathbfcal{P}}\) are updated based on their input tensors \(\mathbf{\Psi}_{t}\) to generate \(\mathbfcal{P}_{t+1}\).
\vspace{2mm}


\textbf{Update} \(\mathbfcal{R}_{t+1}\):
\nolinebreak
\begin{equation}
\begin{aligned} 
\mathbfcal{R}_{t+1}=\argmin_{\mathbfcal{R}}\langle\mathbf{\Gamma}_{t}, \mathbfcal{R} - \mathbfcal{P}_{t+1}\rangle + \frac{\lambda_{3}}{2}\|\mathbfcal{R} - \mathbfcal{Q}_{t}\|_{2}^{2} \\
\end{aligned}
\end{equation}

\begin{equation}
       \Leftrightarrow \mathbfcal{R}_{t+1} = \frac{\mathbfcal{P}_{t+1} + \mathbf{\Gamma}_{t}}{\lambda_{3}}
       \label{eq:rt}
\end{equation} 


\textbf{Update} \(\mathbfcal{Q}_{t+1}\):
\nolinebreak
\begin{equation}
\begin{aligned}
    \mathbfcal{Q}_{t+1} = \argmin_{\mathbfcal{Q}} g_{Q}(\mathbfcal{Q}) + \frac{\lambda_{2}}{2}\|\mathbfcal{Z}_{t} - \mathbfcal{P}_{t+1} \odot \mathbfcal{Q}\|_{2}^{2} \\ + \langle\mathbf{\Omega}_{t}, \mathbfcal{L}_{t}-\mathbfcal{Q}\rangle 
    + \frac{\lambda_{4}}{2}\|\mathbfcal{L}_{t}-\mathbfcal{Q}\|_{2}^{2}
\end{aligned}
\end{equation} 

\begin{equation}
   = \argmin_{\mathbfcal{P}} g_{Q}(\mathbfcal{Q}) + \frac{\mathbf{\mathbfcal{P}_{t+1}^{2}\lambda_{2} + \lambda}_{4}}{2}\Big(\mathbfcal{Q}-\frac{\lambda_{2}\mathbfcal{Z}_{t}\mathbfcal{P}_{t+1} + \lambda_{4}\mathbfcal{L}_{t}+\mathbf{\Omega}_{t}}{\mathbfcal{P}^{2}_{t}\lambda_{2} + \lambda_{4}}\Big)^{2}
\end{equation}

\begin{equation}
   = \mathbfcal{D}_{g_{Q}}(\mathbfcal{P}_{t} - \frac{\mathbf{\Upsilon}_{t}}{\mathbfcal{P}^{2}_{t}\lambda_{2} + \lambda_{4}})
   \label{eq:qt1}
\end{equation} 

where: \(\mathbf{\Upsilon}_{t} = \lambda_{2}\mathbfcal{Z}_{t}\mathbfcal{P}_{t+1} + \lambda_{4}\mathbfcal{L}_{t}+\mathbf{\Omega}_{t}\). Similar to \(\mathbfcal{P}_{t+1}\), we solve (\ref{eq:qt1}) and update \(\mathbfcal{Q}_{t+1}\) using a CNN (\textit{.i.e}, ResUNet \cite{diakogiannis2020resunet}). \\

\vspace{1mm}


\textbf{Update} \(\mathbfcal{L}_{t+1}\):

\begin{equation}
\begin{aligned} 
\mathbfcal{L}_{t+1}=\argmin_{\mathbfcal{L}}\langle\mathbf{\Omega}_{t}, \mathbfcal{L} - \mathbfcal{Q}_{t+1}\rangle + \frac{\lambda_{4}}{2}\|\mathbfcal{L} - \mathbfcal{Q}_{t+1}\|_{2}^{2} \\
\end{aligned}
\end{equation}

\begin{equation}
       \Leftrightarrow \mathbfcal{L}_{t+1} = \frac{\mathbfcal{Q}_{t+1} + \mathbf{\Omega}_{t}}{\lambda_{4}}
       \label{eq:lt}
\end{equation} 
\vspace{2mm}


\textbf{Update} \(\mathbfcal{Z}_{t+1}\):
\nolinebreak
\begin{equation}
\begin{aligned}
    \mathbfcal{Z}_{t+1} = \argmin_{\mathbfcal{Z}}g_{Z}(\mathbfcal{Z}) + \langle\mathbf{\Delta_{t}}, \mathbfcal{I}_{t} - \mathbfcal{Z}\rangle  \\ 
    + \frac{\lambda_{2}}{2} \|\mathbfcal{Z} - \mathbfcal{P}_{t+1}\odot\mathbfcal{Q}_{t+1}\|_{2}^{2} 
    + \frac{\lambda_{5}}{2} \|\mathbfcal{I}_{t} - \mathbfcal{Z}\|_{2}^{2}
\end{aligned}
\end{equation}

\begin{equation}
   = \argmin_{\mathbfcal{\mathbfcal{Z}}}g_{Z}(\mathbfcal{Z}) + \frac{\lambda_{2}+ \lambda_{5}}{2}\Big(\mathbfcal{Z} - \frac{\lambda_{2}\mathbfcal{P}_{t+1}\mathbfcal{Q}_{t+1} + \lambda_{5}\mathbfcal{I}_{t} + \Delta_{t}}{\lambda_{2} + \lambda_{5}}\Big)^{2}
\end{equation} 

\begin{equation}
   = \mathbfcal{D}_{g_{Z}}(\mathbfcal{Z} - \frac{\mathbf{\Pi}_{t}}{\lambda_{2}+\lambda_{5}})
   \label{eq:zt1} 
\end{equation} 

where \(\mathbf{\Pi}_{t} = \lambda_{2}\mathbfcal{P}_{t+1}\mathbfcal{Q}_{t+1} + \lambda_{5}\mathbfcal{I}_{t} + \Delta_{t}\). 

Similar to \(\mathbfcal{P}_{t+1}\), we solve (\ref{eq:zt1}) and update \(\mathbfcal{Z}_{t+1}\) using a CNN (\textit{.i.e}, ResUNet \cite{diakogiannis2020resunet}). \\

\textbf{Update} \(\mathbfcal{I}_{t+1}\):
\nolinebreak
\begin{equation}
\begin{aligned} 
\mathbfcal{I}_{t+1}=\argmin_{\mathbfcal{I}}\frac{\lambda_{1}}{2}\|\mathbfcal{X} - \mathbfcal{H}\otimes\mathbfcal{I}\|_{2}^{2} \\
+ \langle \mathbf{\Delta}, \mathbfcal{I} - \mathbfcal{Z}_{t+1}\rangle + \frac{\lambda_{5}}{2}\|\mathbfcal{I} - \mathbfcal{Z}_{t+1}\|_{2}^{2}
\end{aligned}
\end{equation}

\begin{equation}
       =\argmin_{\mathbfcal{X}}\lambda_{1}(\mathbfcal{H}\mathbfcal{H}^{T}\mathbfcal{I} - \mathbfcal{X}\mathbfcal{H}^{T}) + \lambda_{5}(\mathbfcal{I} - \mathbfcal{Z}_{t+1}) + \mathbf{\Delta}_{t}
       \label{eq:it}
\end{equation} 

To solve Equation (\ref{eq:it}), the Fast Fourier transform (FFT) can
be utilized, and \(\mathbfcal{I}_{t+1}\) can be computed as follow:

\begin{equation}
       \mathbfcal{I}_{t+1} = \mathbfcal{F}^{-1}\left\{\frac{\mathbfcal{F}(\lambda_{1}\mathbfcal{H}^{T}\mathbfcal{X}  + \lambda_{5}\mathbfcal{Z}_{t+1} - \mathbf{\Omega}_{t})}{\lambda_{1}\mathbfcal{F}(\mathbfcal{H})^{2} + \lambda_{5}}\right\}
\end{equation} 

\textbf{Update Multipliers}: Finally, the Lagrange multiplier tensors are
updated following the strategy of the ALM method \cite{lin2010augmented} as

\begin{equation}
   \mathbf{\Gamma}_{t+1} = \mathbf{\Gamma}_{t} + \lambda_{3} * (\mathbfcal{R}_{t+1} - \mathbfcal{P}_{t+1}) 
\end{equation}

\begin{equation}
   \mathbf{\Omega}_{t+1} = \mathbf{\Omega}_{t} + \lambda_{4} * (\mathbfcal{L}_{t+1} - \mathbfcal{Q}_{t+1}) 
\end{equation}

\begin{equation}
   \mathbf{\Delta}_{t+1} = \mathbf{\Delta}_{t} + \lambda_{5} * (\mathbfcal{I}_{t+1} - \mathbfcal{Z}_{t+1}) 
\end{equation}

\subsection{Unrolled Network Architecture}
\label{network}
\begin{figure*}[ht!]
  \centering
   \includegraphics[width=0.72\linewidth]{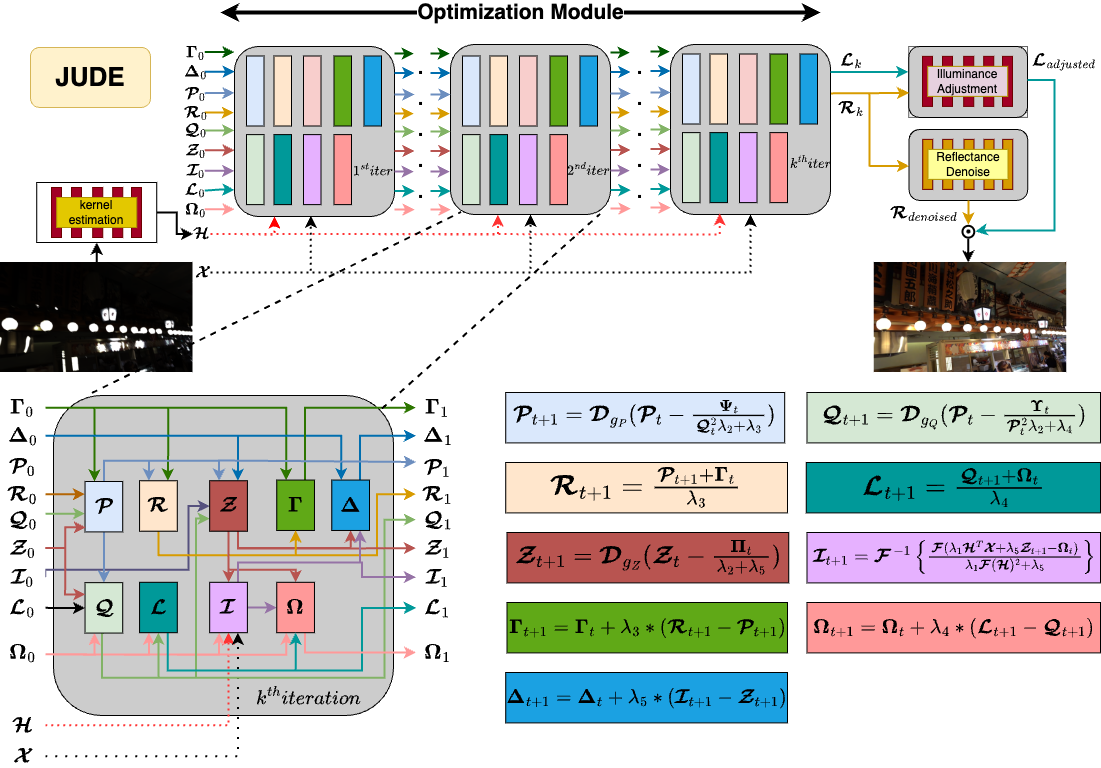}

   \caption{JUDE Architecture. First, all variables and multipliers are initialized with zeros. The input \(\mathbfcal{X}\) is put through a kernel estimation module, and outputs an initial blur kernel \(\mathbfcal{H}\) and a weight map. The proposed JUDE consists of K blocks, wherein each block, \( \mathbfcal{I} \), \( \mathbfcal{R} \), and \( \mathbfcal{L} \) are updated using closed-form solutions, whereas \( \mathbfcal{Z} \), \( \mathbfcal{P} \) and \( \mathbfcal{Q} \) are updated utilizing Convolutional Neural Networks (CNNs), particularly a ResUNet. Decomposed illuminance \( \mathbfcal{L} \) is enhanced, and reflectance \( \mathbfcal{R} \) is denoised before merging into a non-blurry, bright output.}
   \label{fig:jude}
\end{figure*}
\begin{figure*}[ht!]
  \centering
   \includegraphics[width=0.75\linewidth]{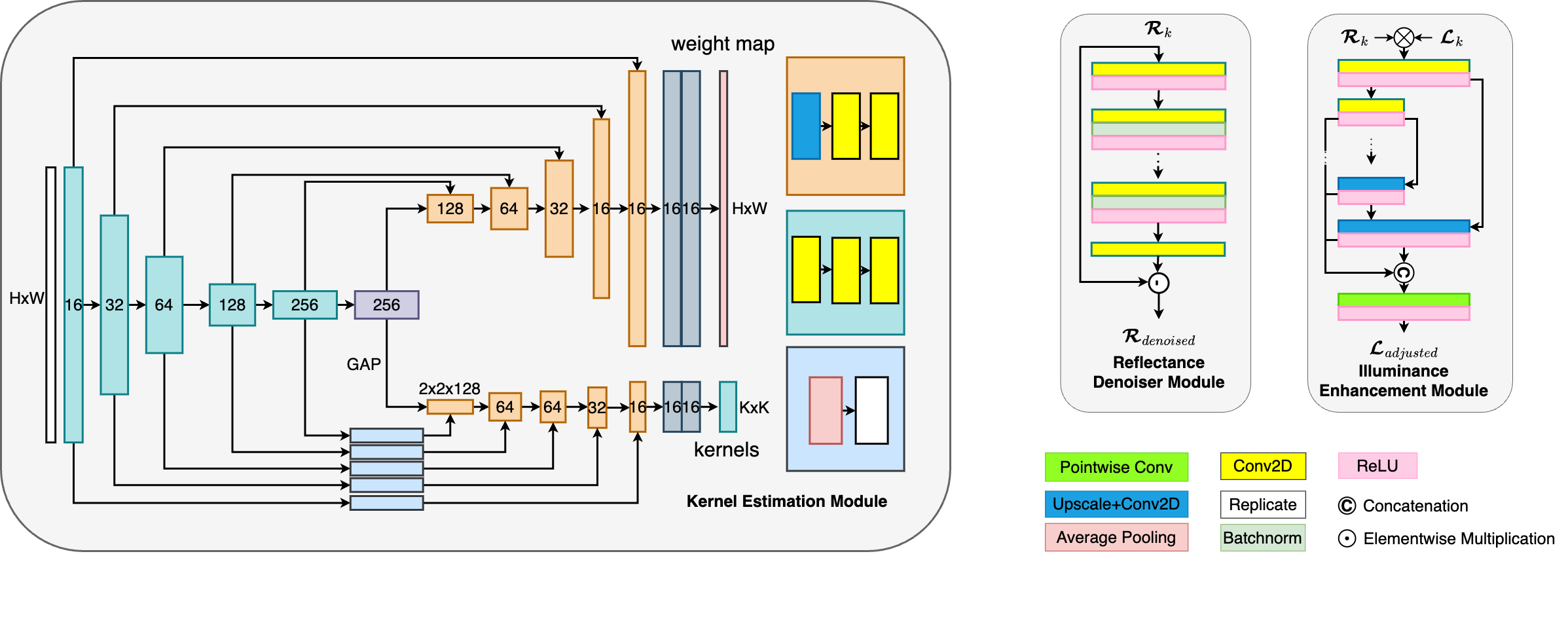}

   \caption{The architecture of the Kernel Estimation (left), Illuminance Enhancement (right), and Reflectance Denoiser (middle) modules.}
   \label{fig:jude2}
\end{figure*}

We address the optimization challenge by transforming the iterative steps into a sequence of blocks, creating a deep unrolling network for low-light image enhancement and deblurring, as shown in Figure \ref{fig:jude}.

\textbf{Kernel Estimation Module.} Initially, a low-light blurry image, \(\mathbfcal{X}\), is processed through a CNN to estimate the blur kernel \(\mathbfcal{H}\). This kernel prediction yields the blur kernel and a pixel-specific weight map to address spatially invariant blur, as shown in Figure \ref{fig:jude2}. Two branches share the same feature extraction part, and then while we add a Decoder module to generate the weight map which has the same size as the input image, the Global Average Pooling technique is adopted to create the initial blur kernel.

\textbf{Optimization Module.} The optimization variables then propagate through $K$ unrolled blocks, with each block's operations mirroring one iteration of the iterative algorithm outlined in Section \ref{optimization}. Updates to \( \mathbfcal{I} \), \( \mathbfcal{R} \), and \( \mathbfcal{L} \) are performed through closed-form solutions, while \( \mathbfcal{P} \), \( \mathbfcal{Q} \), and \( \mathbfcal{Z} \) are updated using CNNs with a ResUNet architecture \cite{diakogiannis2020resunet}. Beyond the network parameters in \(\mathbfcal{D}_{g_{P}}\), \(\mathbfcal{D}_{g_{Q}}\), and \(\mathbfcal{D}_{g_{Z}}\) for \( \mathbfcal{P}_{t} \), \( \mathbfcal{Q}_{t} \), and \(\mathbfcal{Z}_{t}\), other parameters and multipliers are treated as learnable weights, updated through backpropagation during training, including the kernel estimation and hyper-parameters modules. This architecture strictly follows a mathematical model, ensuring complete interpretability and facilitating analysis. 

\begin{table*}[!hb]
\small
\centering
\def\arraystretch{1.15}%
\begin{tabular}{c|c|c|c|c|c}
\hline
\textbf{Category} & \textbf{Methods} & \textbf{Param. (M)} & \textbf{PSNR $\uparrow$} & \textbf{SSIM $\uparrow$} & \textbf{LPIPS $\downarrow$} \\
\hline
\multirow{3}{*}{\textbf{LLE \textrightarrow Deblur}} & FourLLIE \textrightarrow FFTFormer & 16.72 & 18.433 & 0.705 & 0.305 \\
& LLFormer \textrightarrow FFTFormer & 39.12 & 20.290 & 0.792 & 0.212 \\
& RetinexFormer \textrightarrow FFTFormer & 18.21 & 16.452 & 0.702 & 0.324 \\
\hline
\multirow{2}{*}{\textbf{Deblur \textrightarrow LLE}} & MIMO \textrightarrow RetinexFormer & 17.71 & 17.024 & 0.770 & 0.271 \\
 & FFTFormer \textrightarrow RetinexFormer & 18.21 & 16.712 & 0.728 & 0.325 \\
\hline
\multirow{4}{*}{\textbf{LLE \& Deblur}} & FFTFormer & 16.6 & 19.889 & 0.858 & \color{green}{0.139} \\
 & RetinexFormer & 1.61 & 25.505 & \color{green}{0.862} & 0.240 \\
 & LEDNet \cite{lednet} & 7.5 & \color{green}{25.740} & 0.850 & 0.224 \\
 & FELI \cite{feli} & 0.6 & \color{blue}{26.728} & \color{blue}{0.914} & \color{blue}{0.132} \\
 \cline{2-6}
 & \textbf{JUDE (Proposed)} & 35.2 & \color{red}{\textbf{26.884}} & \color{red}{\textbf{0.932}} & \color{red}{\textbf{0.127}} \\
\hline
\end{tabular}
\caption{Quantitative evaluation on the LOL-Blur test set. FFTFormer \cite{fftformer} and RetinexFormer \cite{retinexformer} were retrained on LOL-Blur using publicly available source code. Other model metrics are either acquired from the original paper or from running the provided code. Model parameters are represented in millions (M). Red-bold, blue, and green fonts denote the top three performances, respectively.}
\label{table:1}
\end{table*}

\textbf{Illuminance Enhancement and Reflectance Denoiser Modules.} In the final step, we attach an Illuminance Enhancement Module to enhance the optimized low-light sharp illuminance while putting estimated reflectance through a Denoiser to further remove possible noise before merging and getting the final bright sharp output. The illustration of these two modules is shown in Figure \ref{fig:jude2}. Specifically, the illuminance enhancer module consists of an Encoder and a Decoder with a point-wise convolution to extract features and generate the enhanced luminance map. Additionally, we integrate the skip connection \cite{he2016deep} to preserve useful features during the learning process. Reflectance denoiser follows the concept of IRCNN \cite{ircnn} which estimates the residual map and subtracts it to get the noise-free output.
\vspace{1mm}

\section{Experiments}

\subsection{Datasets and Implementation Details}

{\setlength{\parindent}{0cm}
\textbf{Configuration.} We train our models end-to-end on the synthetic LOL-Blur dataset \cite{lednet}, which includes 10,200 image pairs for training and 1,800 for testing. Using the Adam optimizer with \(\beta_{1} = 0.9\) and \(\beta_{2} = 0.99\), the training process comprises three phases. Initially, we train the model with random crops of \(256 \times 256\) for 300 epochs. Next, we fine-tune the best model from the first phase using random crops of \(384 \times 384\) for 100 epochs, and finally, we refine the model with image patches sized \(512 \times 512\) with random noise added for another 100 epochs. Batch sizes for each phase are 12, 8, and 4, respectively, with initial learning rates of \(1\times10^{-4}\), \(5\times10^{-5}\), and \(2.5\times10^{-5}\). We utilize the Cosine Annealing Scheduler \cite{cosine} across all phases, setting the minimum learning rate to \(1\times10^{-7}\). In our experiments, we set the number of blocks to \(K=5\) and the blur kernel size to \(31\times31\). 

{\setlength{\parindent}{0cm}
\textbf{Loss Fuction.} The loss function used in all phases is as follows:
\begin{equation}
    \mathbfcal{L} = \mathbfcal{L}_{MAE} +  \sigma\mathbfcal{L}_{FFT}
    \label{eq:loss}
\end{equation}

while \(\mathbfcal{L}_{MAE}\) constraints the color and brightness level, \(\mathbfcal{L}_{fft}\) constraints the sharpness of the image. The hyper parameter \(\sigma\) is set to \(0.1\) to control the weight of \(\mathbfcal{L}_{FFT}\).

In details, \(\mathbfcal{L}_{MAE}\) is defined as:
\begin{equation}
    \mathbfcal{L}_{MAE} = ||\hat{\mathbfcal{Y}} - \mathbfcal{Y}||_{1}
\end{equation}

By observing that, to sharpen the output, the difference between the high frequency of output and ground truth should be minimized, therefore \(\mathbfcal{L}_{FFT}\) is called to calculate the Fast Fourier Transform difference between the two, as shown below:
\begin{equation}
    \mathbfcal{L}_{FFT} = ||\mathbfcal{F}(\hat{\mathbfcal{Y}}) - \mathbfcal{F}(\mathbfcal{Y})||_{1}
\end{equation}
where \(\mathbfcal{F}(\hat{\mathbfcal{Y}})\), \(\mathbfcal{F}(\mathbfcal{Y})\) represent the Fast Fourier Transform of the network output and the ground truth, respectively.

\textbf{Evaluation.} We quantitatively and qualitatively compare the proposed JUDE with LEDNet \cite{lednet}, FELI \cite{feli} which are specifically designed for a joint task. We also comprehensively investigate the performance of recent image
processing methods on this task, i.e., low-light enhancement
(LLE) methods: FourLLIE \cite{wang2023fourllie}, LLFormer \cite{wang2023ultra}, and RetinexFormer \cite{retinexformer}
deblurring methods: MIMO \cite{Cho2021RethinkingCA}, FFTFormer \cite{fftformer}, and FFTformer \cite{fftformer}. We also retrained FFTFormer \cite{fftformer} and RetinexFormer \cite{retinexformer} using LOL-Blur dataset with the publicly provided code by the author and report in this paper. We combine and validate the performance of JUDE against others in three categories:

{\setlength{\parindent}{0.5cm}
\textbf{Enhancement \textrightarrow  Deblurring:} We select FourLLIE \cite{wang2023fourllie}, LLFormer \cite{wang2023ultra}, and RetinexFormer \cite{retinexformer} for enhancement, followed by FFTFormer \cite{fftformer} for deblurring.

{\setlength{\parindent}{0.5cm}
\textbf{Deblurring \textrightarrow  Enhancement:} For deblurring, we use MIMO \cite{Cho2021RethinkingCA}, and FFTFormer \cite{fftformer}. For low-light enhancement, we choose RetinexFormer \cite{retinexformer}, the state-of-the-art in this task.

{\setlength{\parindent}{0.5cm}
\textbf{End-to-end training on LOL-Blur Dataset:} We conduct end-to-end training on the LOL-Blur dataset by retraining several state-of-the-art baselines, including RetinexFormer \cite{retinexformer} and FFTFormer \cite{fftformer}, state-of-the-arts in LLE and deblurring using their available code.

\subsection{LOL-Blur Dataset Evaluation}

{\setlength{\parindent}{0cm}
\textbf{Evaluation Metrics.} We use PSNR and SSIM metrics to assess the performance on the synthetic LOL-Blur dataset. Additionally, for evaluating the perceptual quality of the restored images, we include the LPIPS \cite{lpips} metric as a reference. \\

\begin{figure*}[!ht]
    \centering
    \begin{subfigure}[b]{0.245\textwidth}
        \includegraphics[width=\textwidth]{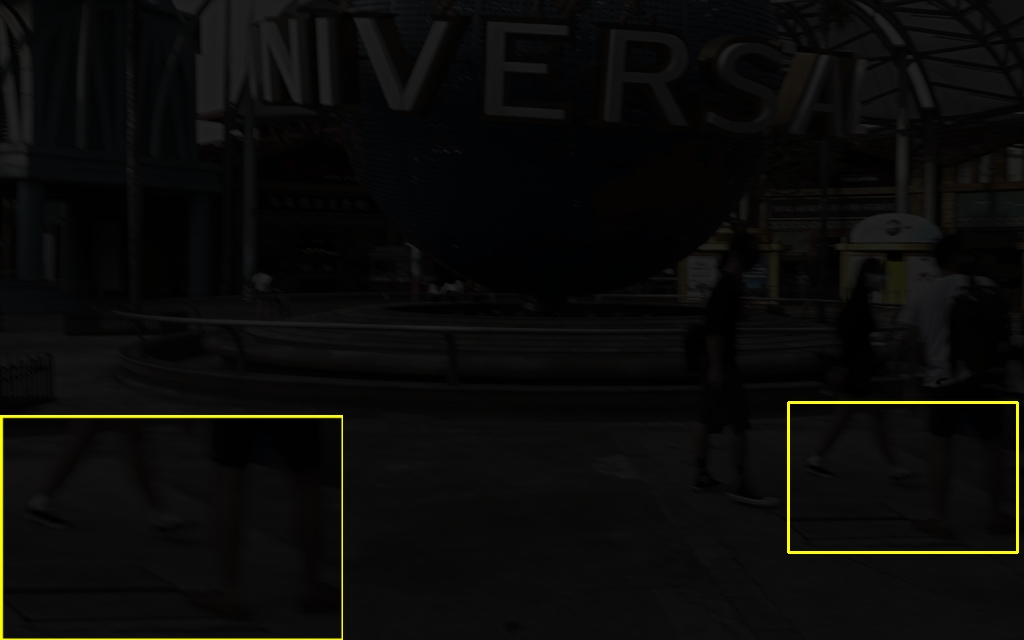}
        \caption{Input}
    \end{subfigure}
    \hfill
    \begin{subfigure}[b]{0.245\textwidth}
        \includegraphics[width=\textwidth]{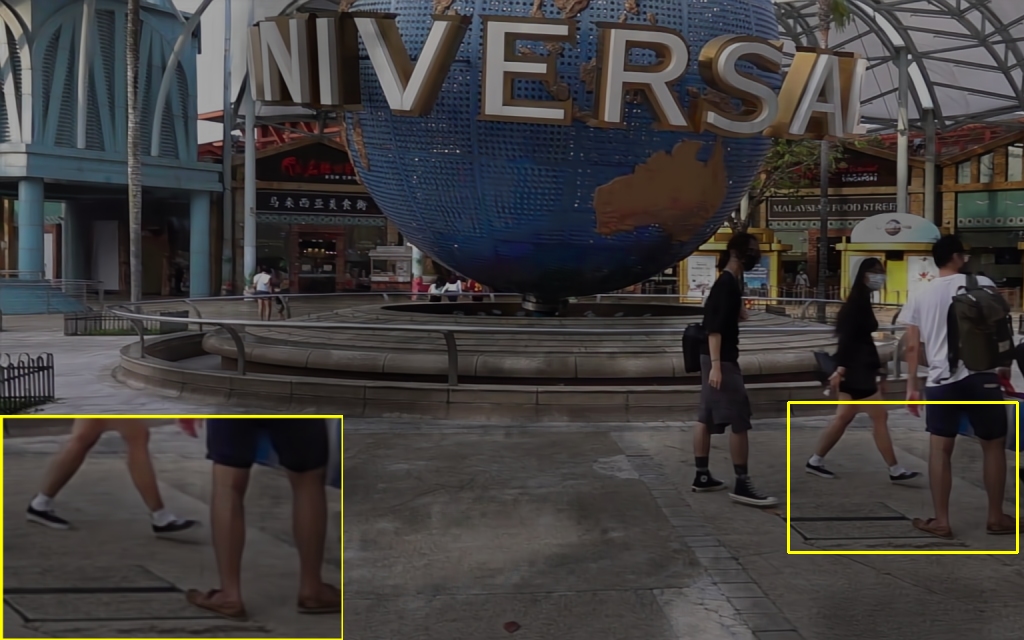}
        \caption{FourLLIE →FFTFormer}
    \end{subfigure}
    \hfill
    \begin{subfigure}[b]{0.245\textwidth}
        \includegraphics[width=\textwidth]{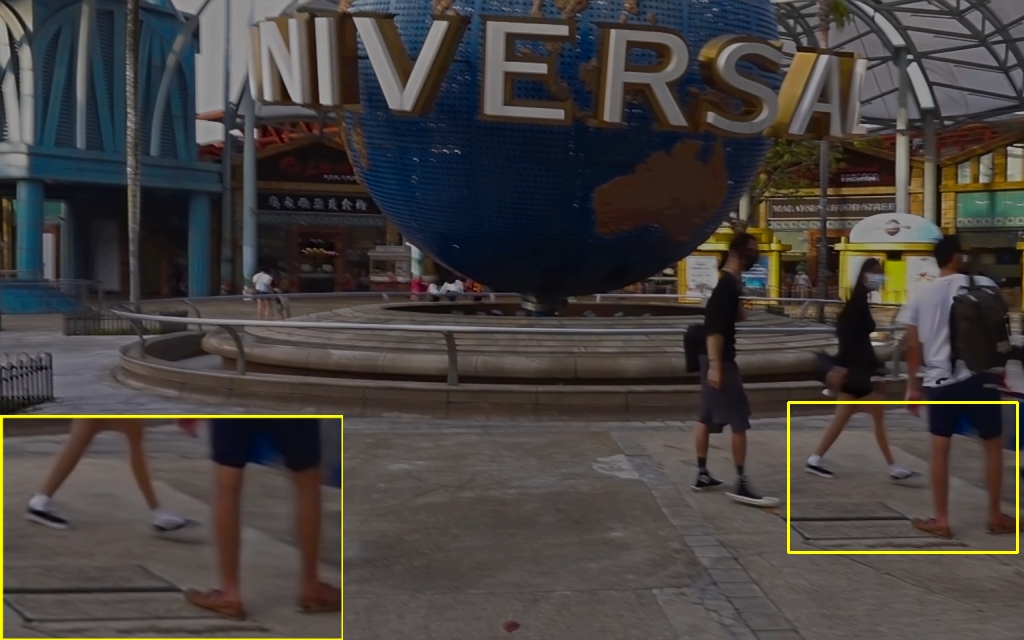}
        \caption{LLFormer →FFTFormer}
    \end{subfigure}
    \hfill
    \begin{subfigure}[b]{0.245\textwidth}
        \includegraphics[width=\textwidth]{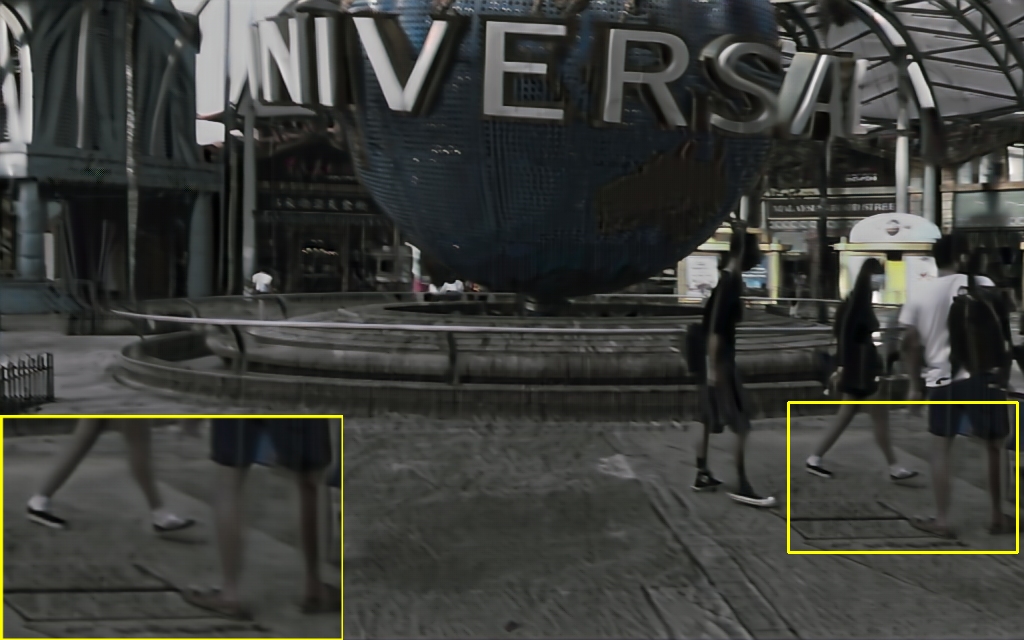}
        \caption{RetinexFormer →FFTFormer}
    \end{subfigure}
    
    \smallskip 
    
    \begin{subfigure}[b]{0.245\textwidth}
        \includegraphics[width=\textwidth]{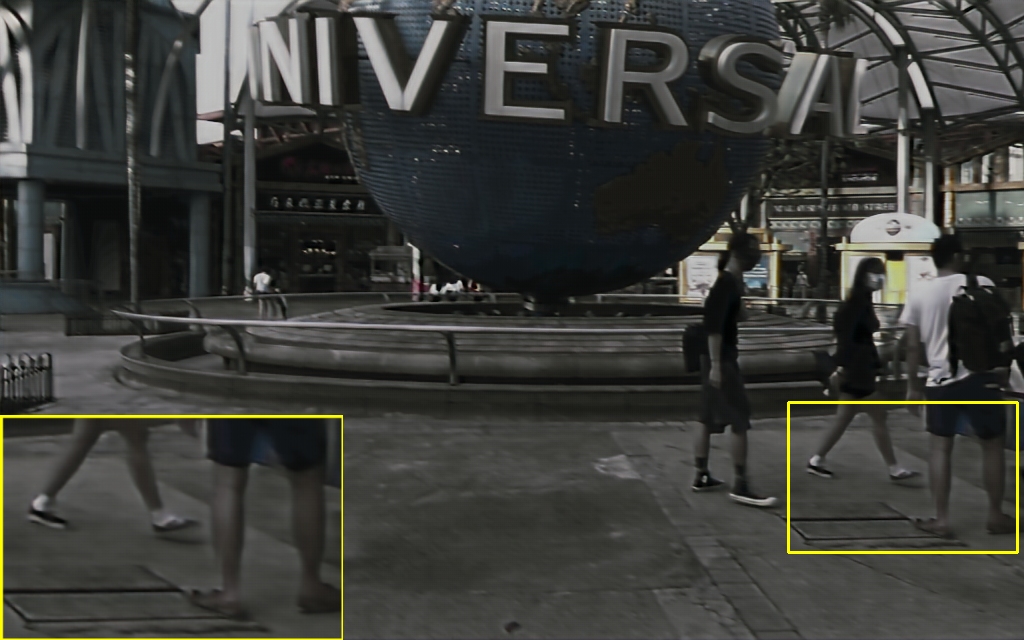}
        \caption{MIMO →RetinexFormer}
    \end{subfigure}
    \hfill
    \begin{subfigure}[b]{0.245\textwidth}
        \includegraphics[width=\textwidth]{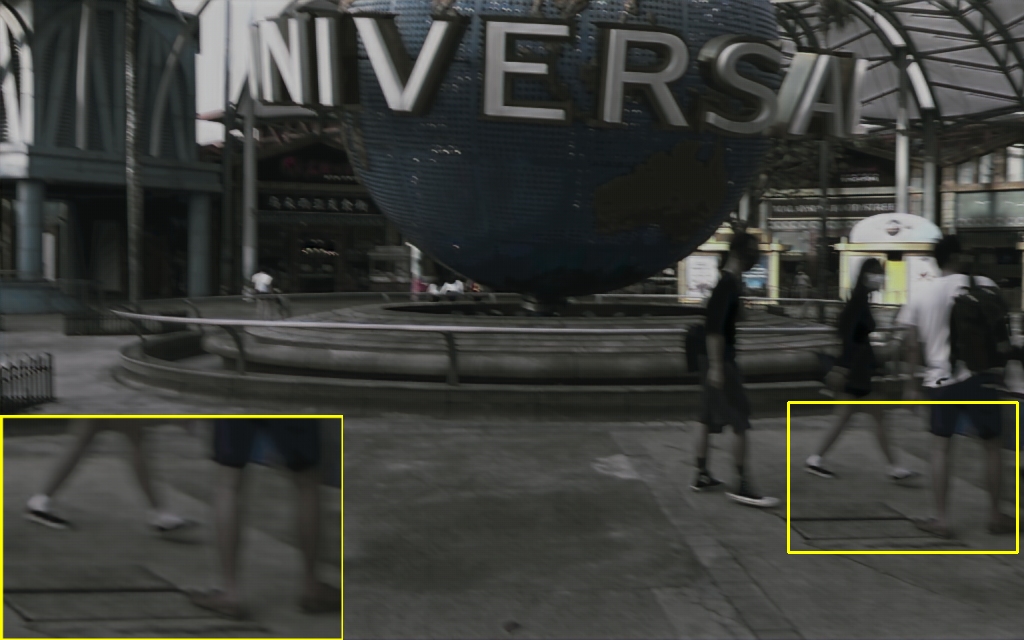}
        \caption{FFTFormer →RetinexFormer}
    \end{subfigure}
    \hfill
    \begin{subfigure}[b]{0.245\textwidth}
        \includegraphics[width=\textwidth]{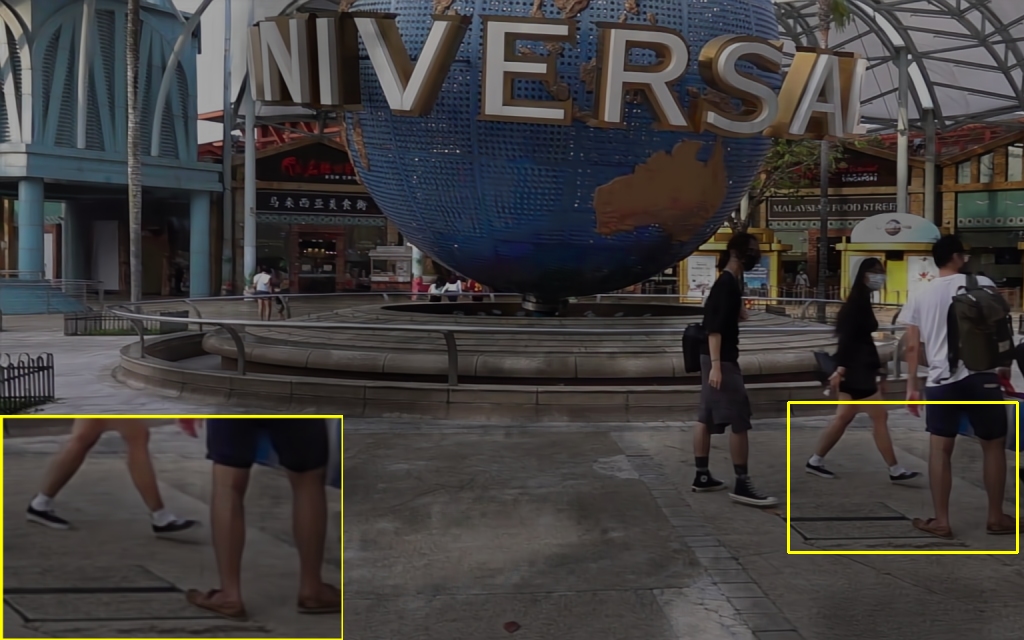}
        \caption{FFTFormer}
    \end{subfigure}
    \hfill
    \begin{subfigure}[b]{0.245\textwidth}
        \includegraphics[width=\textwidth]{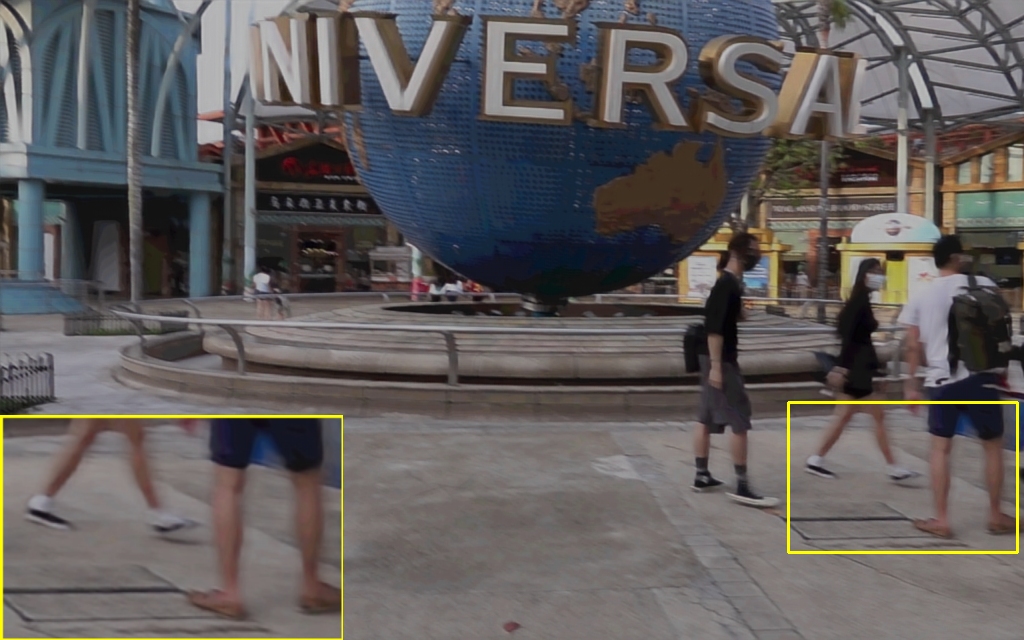}
        \caption{RetinexFormer}
    \end{subfigure}
    
    \smallskip 
    
    \begin{subfigure}[b]{0.245\textwidth}
        \includegraphics[width=\textwidth]{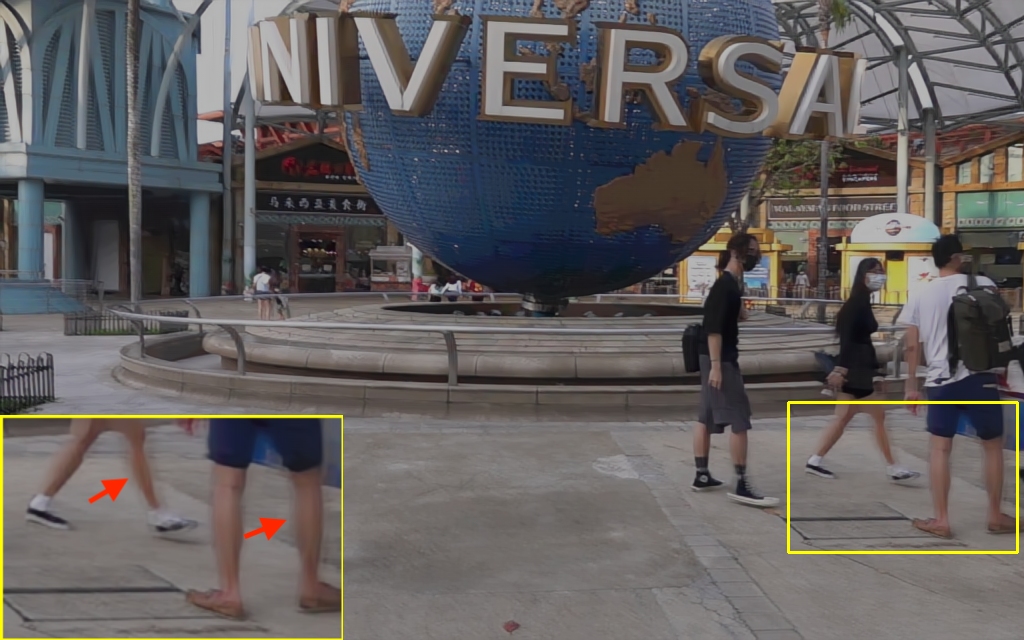}
        \caption{LEDNet}
    \end{subfigure}
    \hfill
    \begin{subfigure}[b]{0.245\textwidth}
        \includegraphics[width=\textwidth]{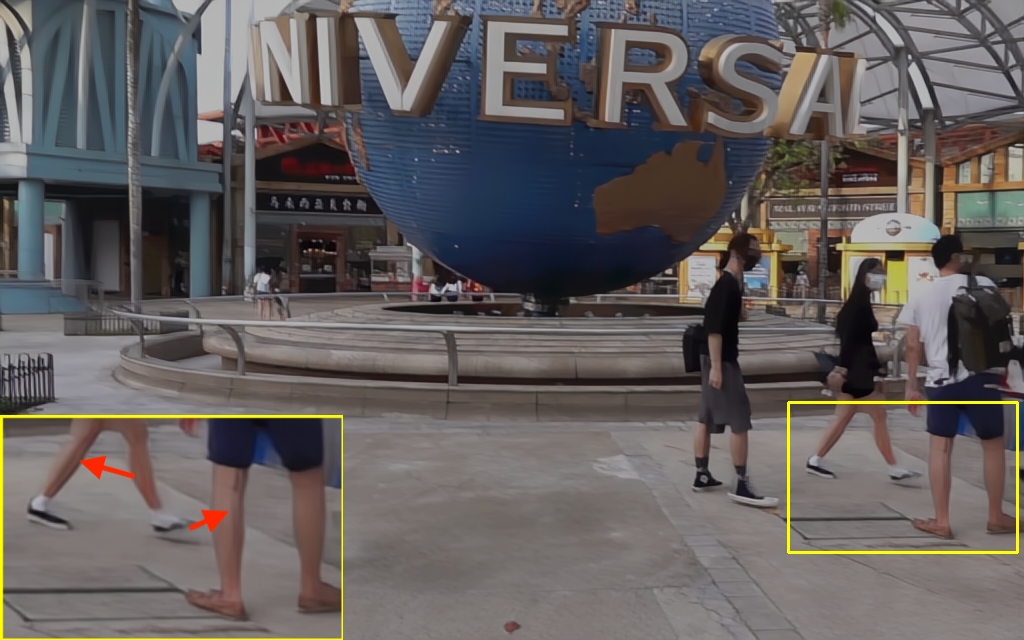}
        \caption{FELI}
    \end{subfigure}
    \hfill
    \begin{subfigure}[b]{0.245\textwidth}
        \includegraphics[width=\textwidth]{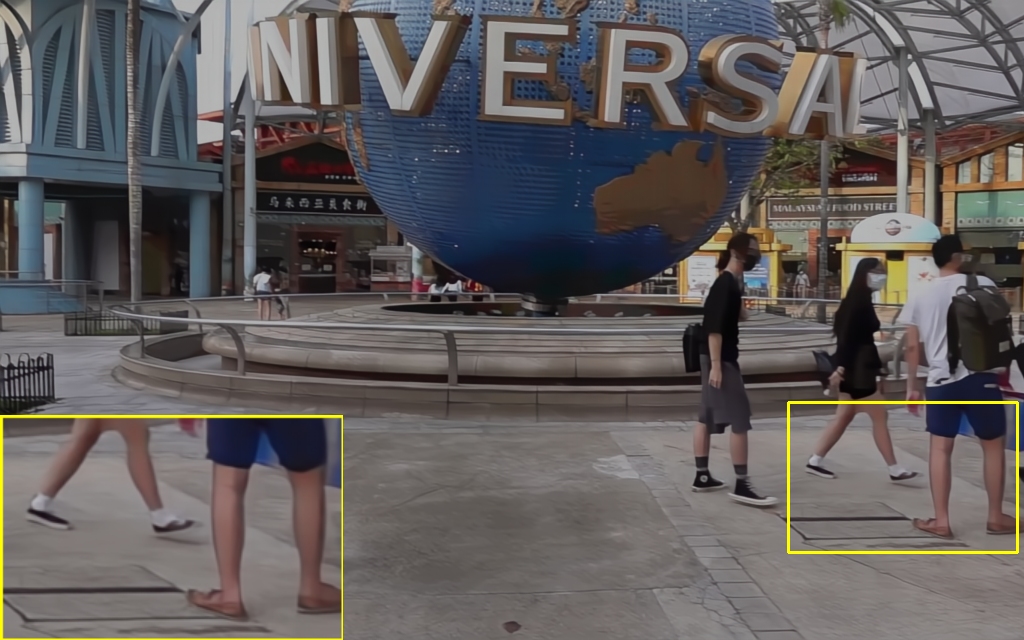}
        \caption{JUDE (Ours)}
    \end{subfigure}
    \hfill
    \begin{subfigure}[b]{0.245\textwidth}
        \includegraphics[width=\textwidth]{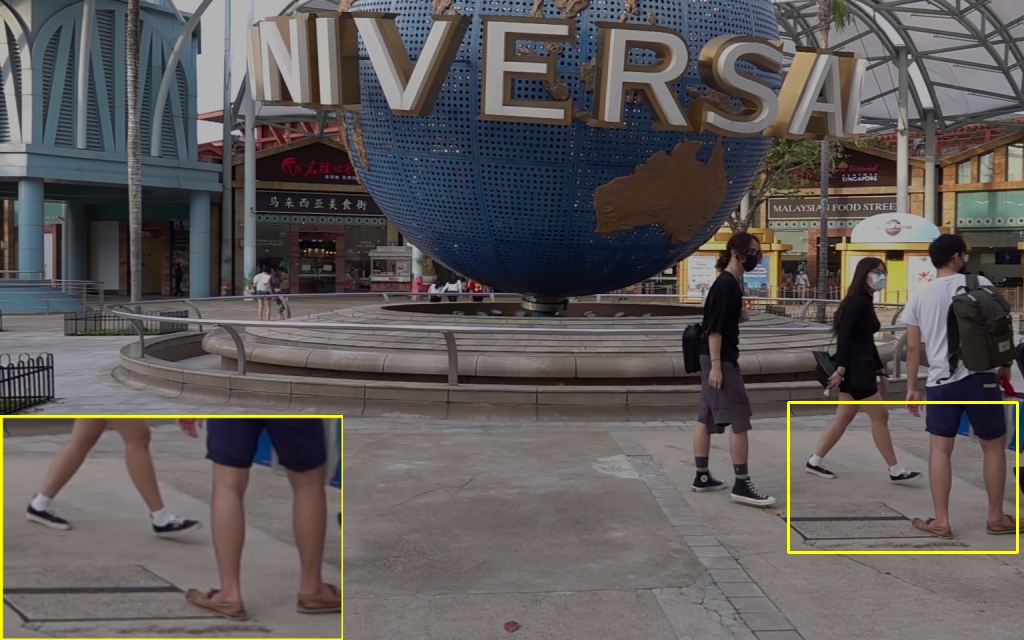}
        \caption{GT}
    \end{subfigure}
    
    \caption{Visual comparison on LOL-Blur dataset. The yellow box indicates obvious differences and is shown at the bottom-left of each result image. (Zoom in for better visualization.)}
    \label{fig:images_LOL-Blur}
\end{figure*}
{\setlength{\parindent}{0cm}
\textbf{Quantitative Evaluation:} Table \ref{table:1} illustrates that JUDE significantly outperforms all other methods in all metrics scores, underscoring its significant deblurring performance and low-light image enhancement. 

\begin{table*}[!t]
\small
\centering
\def\arraystretch{1.15}%
\begin{tabular}{c|c|c|c|c|c|c|c}
\hline
\textbf{Category} & \textbf{Methods} & \textbf{ARNIQA$\uparrow$} & \textbf{CONTRIQUE $\uparrow$} & \textbf{LIQE $\uparrow$} & \textbf{MUSIQ $\uparrow$} & \textbf{CLIPIQA $\uparrow$} & \textbf{DBCNN $\uparrow$} \\
\hline
\multirow{3}{*}{\textbf{LLE \textrightarrow Deblur}} & FourLLIE \textrightarrow FFTFormer & 0.377 & 46.823 & 1.113 & 30.840 & 0.217 & 0.261\\
& LLFormer \textrightarrow FFTFormer & 0.401 & 44.743 & 1.158 & \color{green}{36.534} & 0.208 & 0.257\\
& RetinexFormer \textrightarrow FFTFormer & 0.364 & 41.495 & 1.075 & 34.793 & 0.227 & 0.279\\
\hline
\multirow{2}{*}{\textbf{Deblur \textrightarrow LLE}} & MIMO \textrightarrow RetinexFormer & 0.413 & 40.773 & 1.137 & 33.242 & 0.207 & 0.276\\
 & FFTFormer \textrightarrow RetinexFormer & 0.405 & \color{green}{48.814} & 1.195 & 35.511 & 0.221 & 0.303\\
\hline
\multirow{4}{*}{\textbf{LLE \& Deblur}} & FFTFormer & 0.402 & 38.005 & 1.141 & 32.079 & \color{blue}{0.289} & \color{blue}{0.307}\\
 & RetinexFormer & 0.418 & 43.410 & 1.074 & 31.782 & 0.187 & 0.232\\
 & LEDNet \cite{lednet} & \color{green}{0.419} & \color{blue}{49.828} & \color{blue}{1.414} & \color{blue}{43.623} & \color{green}{0.281} & \color{green}{0.306}\\
 & FELI \cite{feli} & \color{blue}{0.429} & 42.354 & \color{green}{1.155} & 33.669 & 0.207 & 0.239\\
 \cline{2-8}
 & \textbf{JUDE (Proposed)} & \color{red}{\textbf{0.437}} & \color{red}{\textbf{50.207}} & \color{red}{\textbf{1.454}} & \color{red}{\textbf{44.732}} & \color{red}{\textbf{0.299}} & \color{red}{\textbf{0.313}}\ \\
 \hline
\end{tabular}
\caption{Quantitative evaluation on the Real-LOL-Blur data set. FFTFormer \cite{fftformer} and RetinexFormer \cite{retinexformer} were retrained on LOL-Blur using publicly available source code. Red-bold, blue, and green fonts highlight the top three performances, respectively.}
\label{table:2}
\end{table*}

\textbf{Qualitative Evaluation:} Figure \ref{fig:images_LOL-Blur} demonstrates the efficacy of JUDE compared to other methods. Specifically, combinations of separate low-light enhancement and deblurring methods yield unsatisfactory results in terms of color fidelity and detail preservation. RetinexFormer \cite{retinexformer}, FFTFormer \cite{fftformer}, LEDNet \cite{lednet} and FELI \cite{feli} exhibit noticeable distortions and struggle to reconstruct details in areas with significant motion, marked in the yellow rectangle in Figure \ref{fig:images_LOL-Blur}, even after being retrained/trained on the LOL-Blur dataset \cite{lednet}. In contrast, our proposed JUDE faithfully recovers details and enhances the image to a satisfactory level.

\subsection{Real-World Data Evaluation}
The Real-LOL-Blur dataset \cite{rim2020real} comprises 1,354 night-time blurry images sourced from real-world scenarios. 


\begin{figure*}[!ht]
\centering
\subfloat[Input]{\includegraphics[scale=.11]{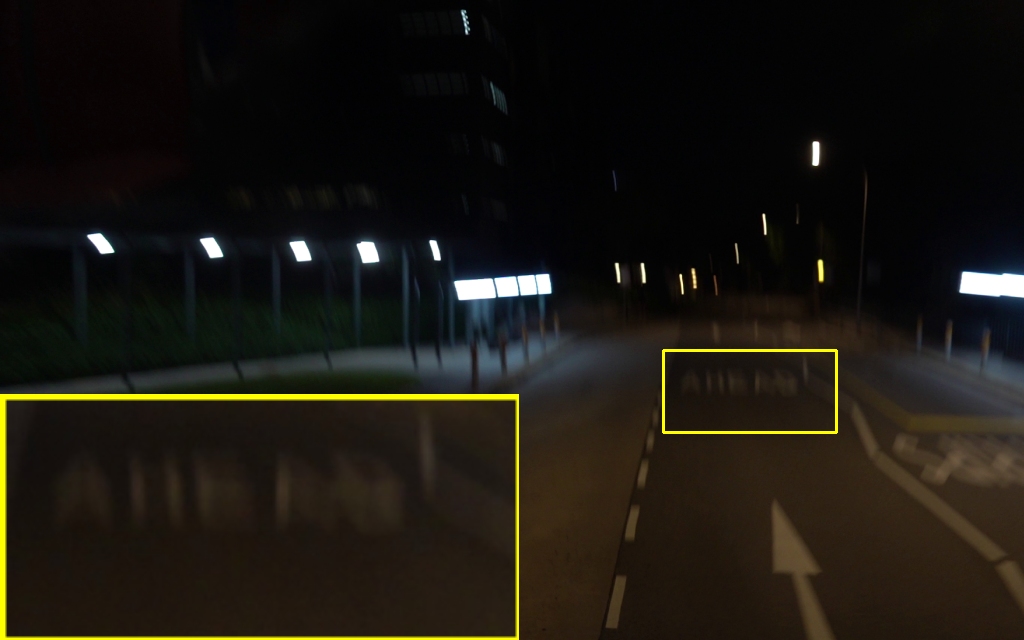}}\,
\subfloat[FourLLIE → FFTFormer]{\includegraphics[scale=.11]{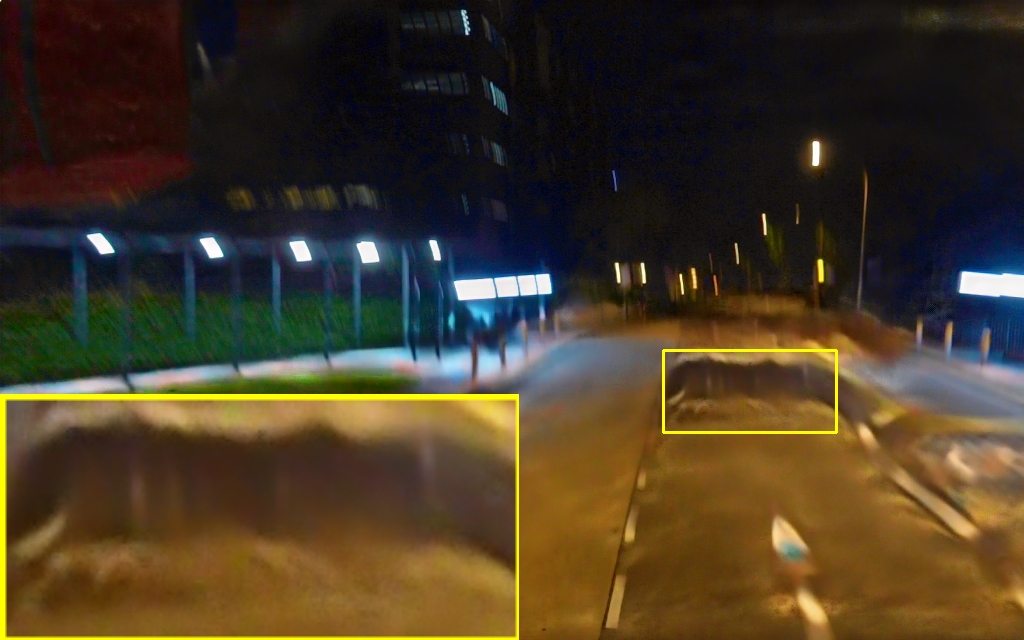}}\,
\subfloat[LLFormer → FFTFormer]{\includegraphics[scale=.11]{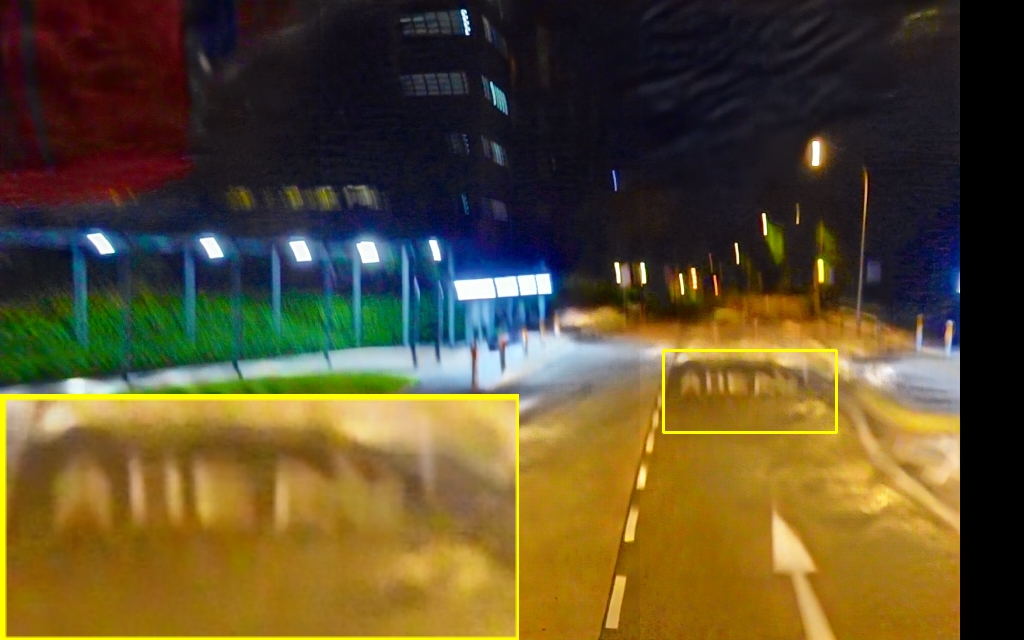}}\,
\subfloat[RetinexFormer → FFTFormer]{\includegraphics[scale=.11]{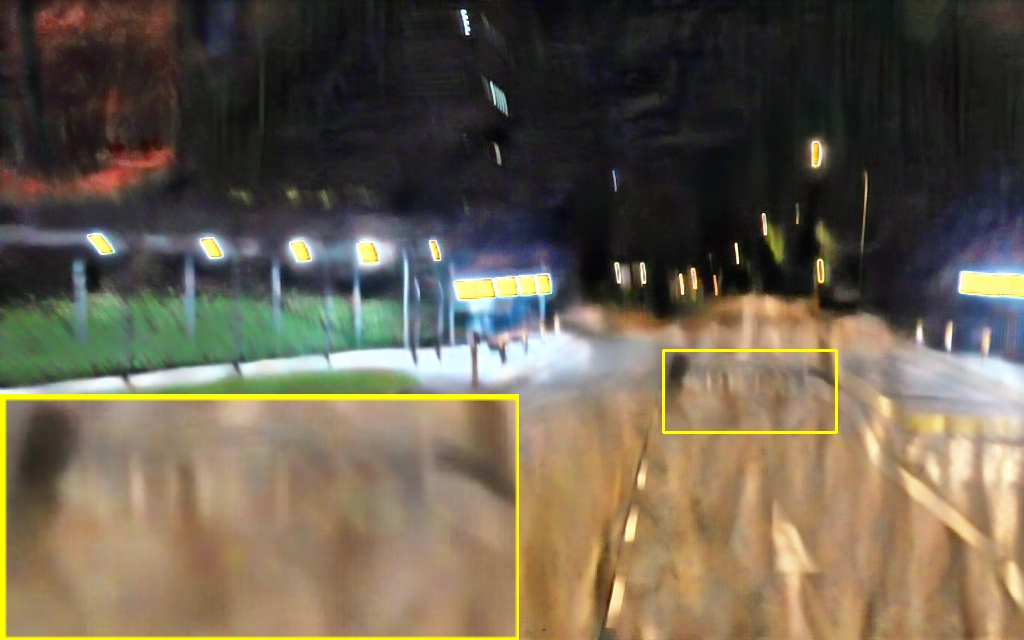}},
\hfill
\subfloat[MIMO → RetinexFormer]{\includegraphics[scale=.11]{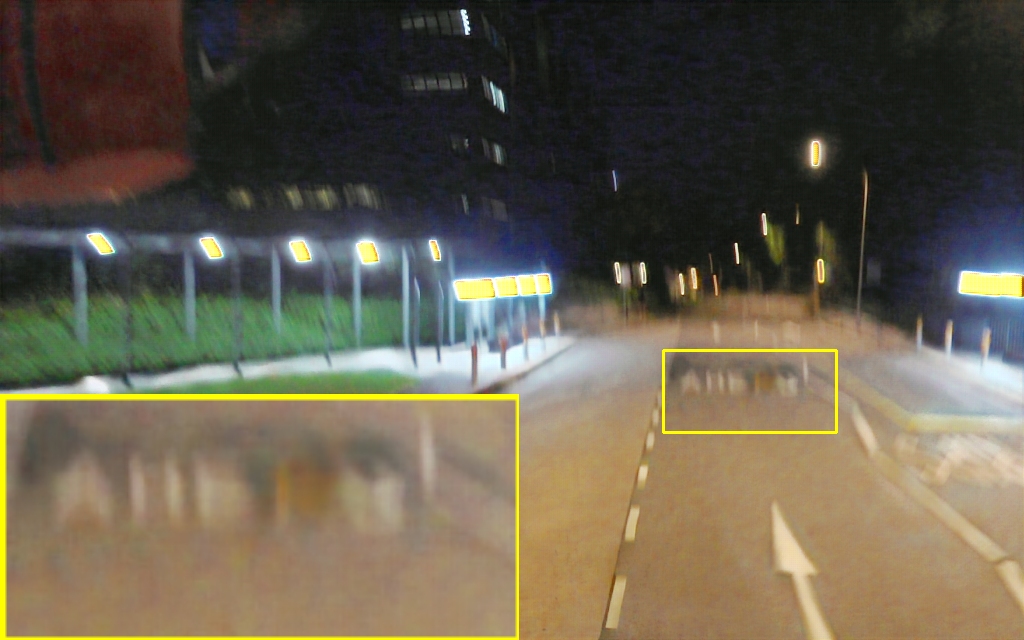}}\,
\subfloat[FFTFormer → RetinexFormer]{\includegraphics[scale=.11]{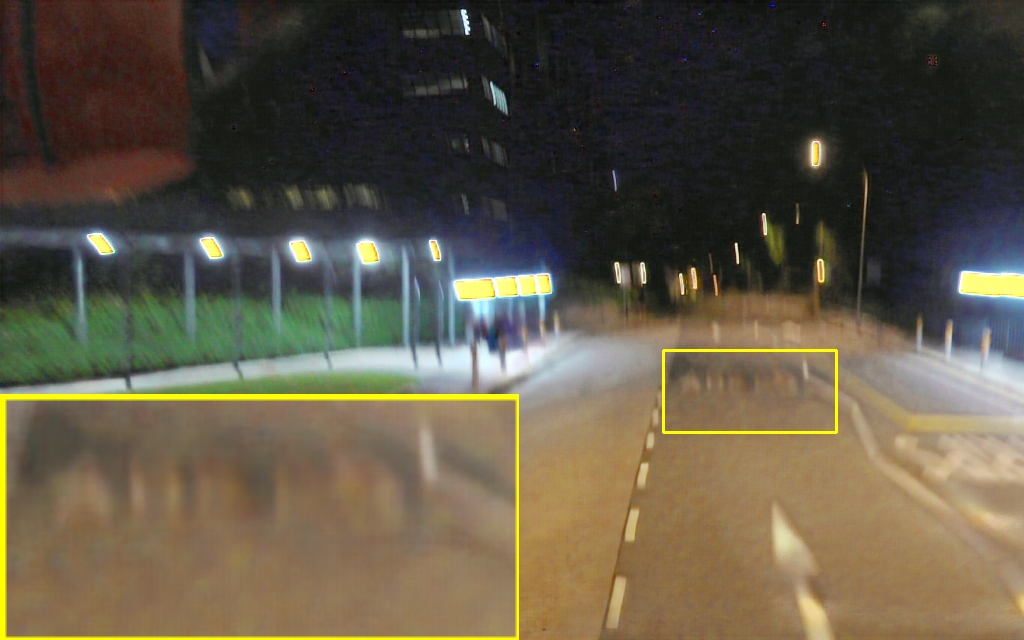}}\,
\subfloat[FFTFormer]{\includegraphics[scale=.11]{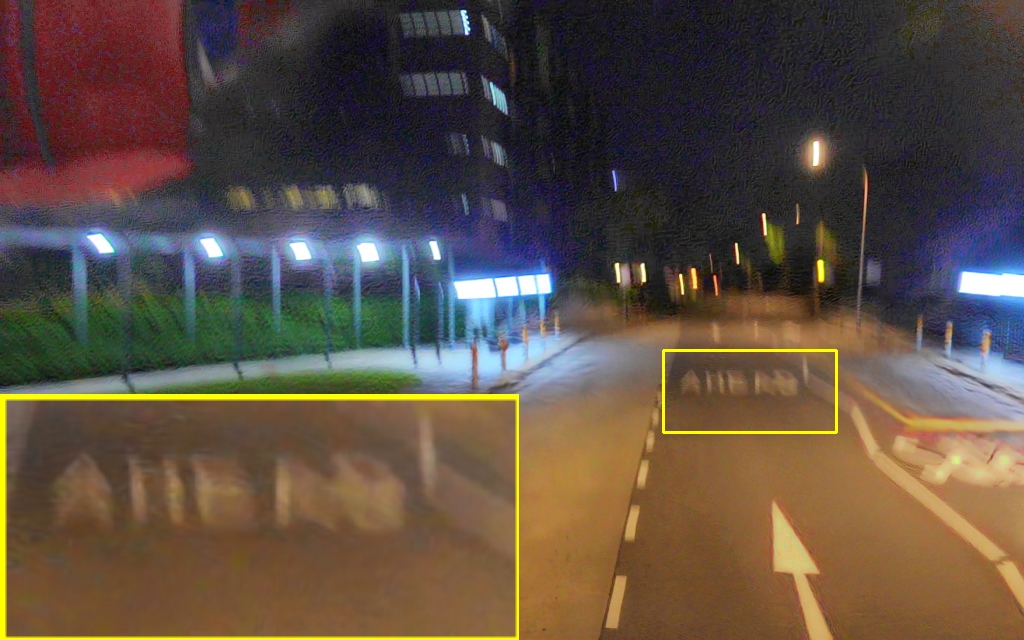}}\,
\subfloat[RetinexFormer]{\includegraphics[scale=.11]{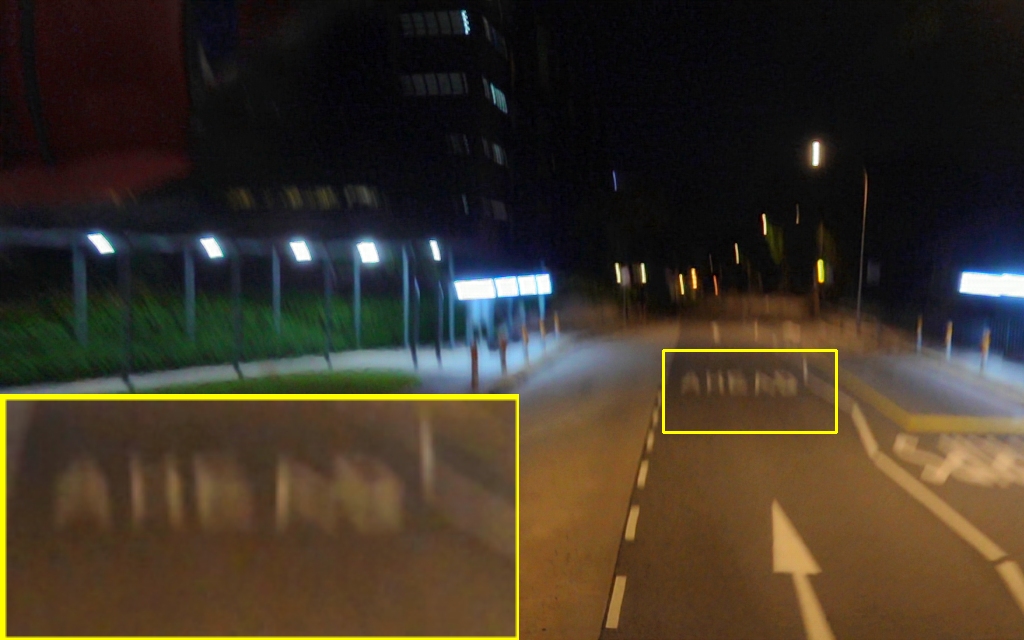}},
\hfill
\subfloat[LEDNet]{\includegraphics[scale=.11]{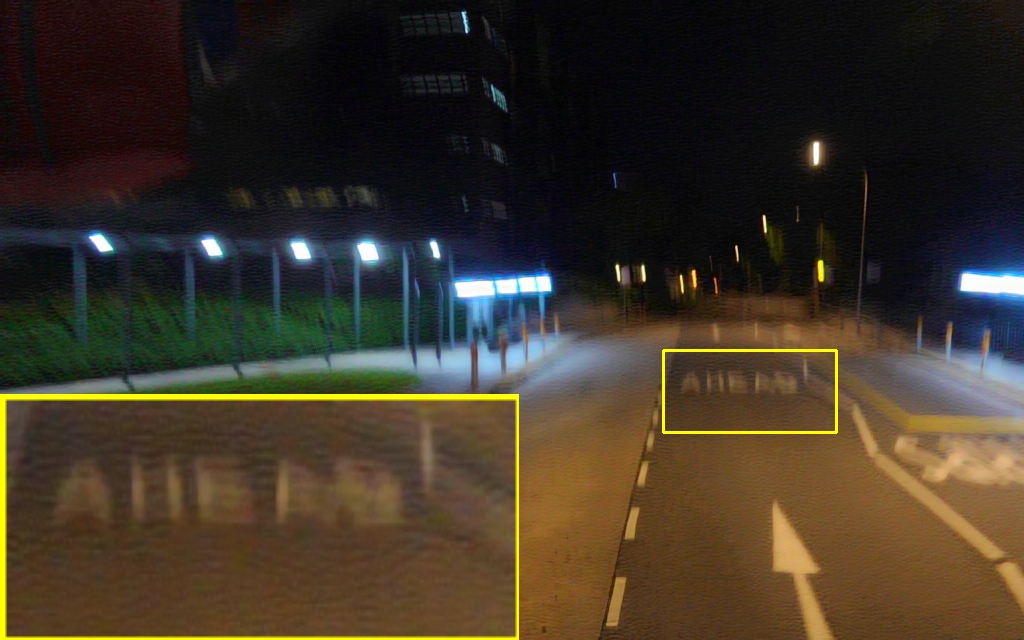}}\,
\subfloat[FELI]{\includegraphics[scale=.11]{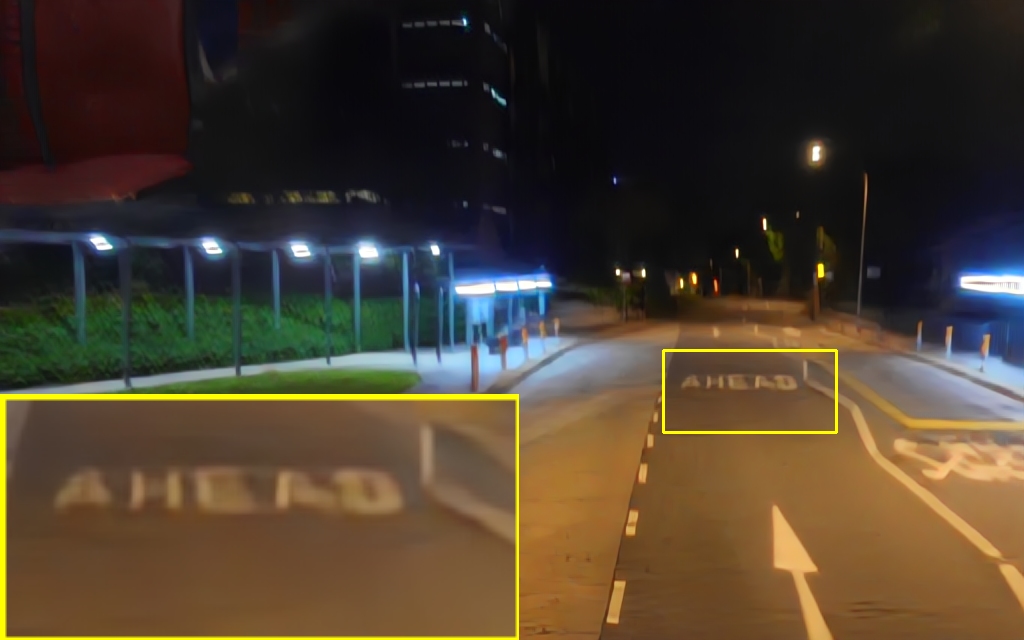}}\,
\subfloat[JUDE (Ours)]{\includegraphics[scale=.11]{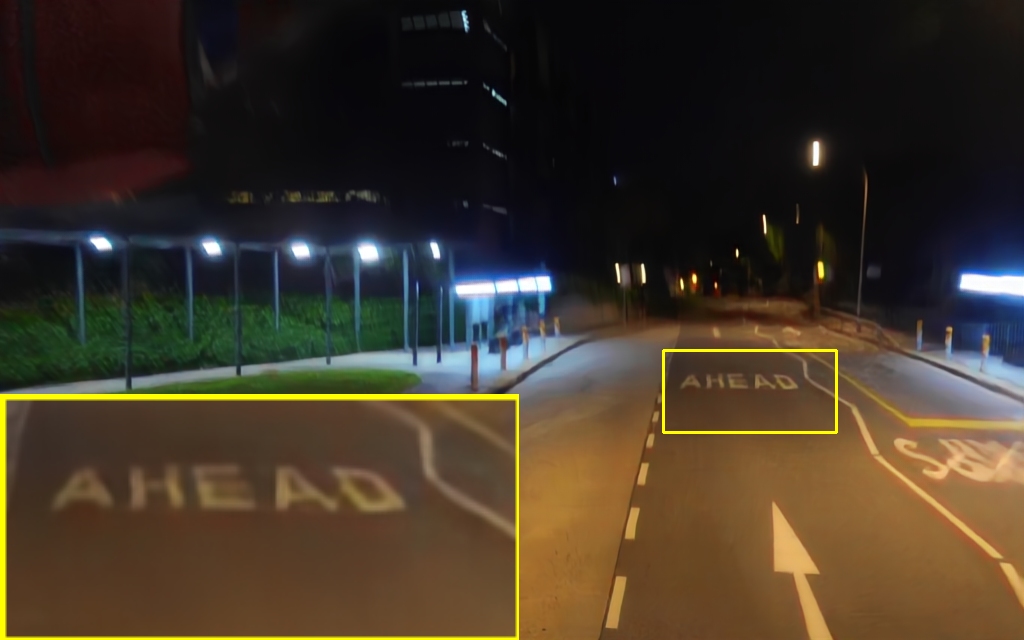}},
\caption{Visual comparison on Real-LOL-Blur \cite{rim2020real} dataset. The yellow box indicates obvious differences and is shown at the bottom-left of each result image. (Zoom in for better visualization.)}
\label{fig:Real-LOL-Blur}
\end{figure*}

\vspace{1mm}
{\setlength{\parindent}{0cm}
\textbf{Evaluation Metrics.} Due to the nature of real-world data, which often includes low-quality images for testing purposes, we utilize the most recent state-of-the-art no-reference image quality assessment (IQA) methods for evaluation: ARNIQA \cite{agnolucci2024arniqa}, CONTRIQUE \cite{madhusudana2022image}, MUSIQ \cite{ke2021musiq}, DBCNN \cite{zhang2020blind}, CLIPIQA \cite{wang2022exploring}, and LIQE \cite{zhang2023blind}. Specifically, ARNIQA leverages deep neural networks for no-reference quality evaluation, while CONTRIQUE employs contrastive learning to enhance representation learning in IQA. MUSIQ assesses multi-scale image quality using transformers, DBCNN utilizes a double-ended CNN architecture for evaluating both distortion and content, CLIPIQA adapts the CLIP \cite{radford2021learning} vision-language model for IQA, and LIQE introduces a learning-based approach that incorporates intrinsic image features for quality assessment. Higher values across these metrics indicate higher image quality.
\vspace{1mm}

{\setlength{\parindent}{0cm}
\textbf{Quantitative  Evaluation.} We present the subjective metrics in Table \ref{table:2}. The result states that the proposed JUDE outperforms all other methods, which indicates that our results are perceptually best in terms of color contrast and sharpness. These metrics also highlight the model’s strong generalization capability, as it is grounded in the physical modeling of low-light blurry images.

\vspace{1mm}
{\setlength{\parindent}{0cm}
\textbf{ Qualitative Evaluation.} Figure \ref{fig:Real-LOL-Blur} provides a comparative analysis of JUDE's performance with other methods on the Real-LOL-Blur dataset \cite{rim2020real}. Various combinations of low-light enhancement and deblurring techniques consistently produce images that are either color-distorted or remain blurry. In addition, methods that have been trained or retrained on the LOL-Blur dataset \cite{lednet} particularly struggle with generalization, failing to accurately reconstruct details and do not faithfully represent the original scenes. In contrast, JUDE demonstrates robust generalization capabilities to effectively recovers intricate details, and enhances the images to a visually pleasing light level even in challenging real-world conditions.


\vspace{-2mm}



\section{Conclusion}
\vspace{-2mm}
We introduced JUDE, a deep joint unfolding network for low-light image enhancement and deblurring, built on a physical formulation of low-light blurry images. JUDE combines model-based interpretability with deep learning generalization ability, integrating data-driven priors into its optimization module. Both quantitative metrics and qualitative results demonstrate JUDE's superior effectiveness in handling synthetic and real-world datasets compared to LEDNet \cite{lednet}, FELI \cite{feli}, and recent low-light enhancement and deblurring techniques, establishing its potential as a robust solution for enhancing low-light and blurred images. Additionally, the architecture of JUDE is designed to adapt to varying degrees of blur and illumination, enhancing its versatility across different blurring conditions. 

Although our model achieved state-of-the-art results on established datasets, its parameter count imposes certain limitations. This constraint suggests room for improvement, particularly by exploring alternative kernel prediction models and/or substituting the CNNs within the data module with more efficient architectures without compromising its performance. We recognize these challenges and aim to address them in our future work to further enhance the model's performance and efficiency.



{\small
\bibliographystyle{ieee_fullname}
\bibliography{wacv}
}

\end{document}